\documentclass[3p,times]{elsarticle}

\usepackage{amssymb}

\usepackage[figuresright]{rotating}

\begin{document}

\begin{frontmatter}




\title{A Survey of Advanced Computer Vision Techniques for Sports}


\author[label1,label2]{Tiago Mendes-Neves}
\author[label1,label3]{Luís Meireles}
\author[label1,label2]{João Mendes-Moreira}

\address[label1]{Faculdade de Engenharia, Universidade do Porto, Porto, Portugal}
\address[label2]{LIAAD - INESC TEC, Porto, Portugal}
\address[label3]{FC Porto – Futebol SAD, Porto, Portugal}

\begin{abstract}
Computer Vision developments are enabling significant advances in many fields, including sports. Many applications built on top of Computer Vision technologies, such as tracking data, are nowadays essential for every top-level analyst, coach, and even player.
In this paper, we survey Computer Vision techniques that can help many sports-related studies gather vast amounts of data, such as Object Detection and Pose Estimation. We provide a use case for such data: building a model for shot speed estimation with pose data obtained using only Computer Vision models. Our model achieves a correlation of 67\%. 
The possibility of estimating shot speeds enables much deeper studies about enabling the creation of new metrics and recommendation systems that will help athletes improve their performance, in any sport. The proposed methodology is easily replicable for many technical movements and is only limited by the availability of video data.

\end{abstract}


\end{frontmatter}


\section{Introduction}
\label{sec:intro}

Most data used in sports comes from either sensors or manual annotations.
Using sensors to obtain data points in sports has several drawbacks, with the most severe being the negative impact on player performance. Furthermore, many sensors are not authorized to be used in competitive games. For most use cases, acquiring data requires non-intrusive technologies - for example, manual annotations. However, manual annotations are very expensive to produce and prone to errors. Manually acquired data also reduces the possibility of systematic analysis in many problems and is impossible for others such as biomechanical analysis.
We believe Computer Vision can have an important role in improving the quality and quantity of data available in sports, both from competitive games and training drills. 

One of the main goals of Computer Vision is to extract information from images. The recent link between Computer Vision and Deep Learning has led to remarkably high efficiency in some of the hardest to execute tasks: classification (i.e., what is the object in the image), detection (i.e., what is the object and where is it located in the image), and detection of specific patterns (e.g., pose/gait estimation). 

The extraction of information from images is essential for many use cases. Autonomous vehicles, which need incredibly high accuracy for the detection of road elements, can save thousands of lives on the road and allow users to be productive during commuting time. Agriculture, for increased automation, will decrease the cost of goods, allowing us to feed a growing population. And for healthcare, for automated, more efficient, and convenient disease detection, saving thousands of patients from the problems related to late diagnostics. There are many resources being invested into Computer Vision and with good reason.

In this article, we present a detailed but accessible review of the state-of-the-art methods for Object Detection and Pose Estimation from video, two tasks with the potential to acquire a tremendous amount of valuable data automatically. Data collected will serve as a basis for advanced analytic systems, which today seem out of reach. We present single-view methods that only rely on capturing video from one angle, which has similar performance to multi-view methods while being much simpler to use. 

Current commercial data collection solutions fall into three categories: (1) collection of sensor data (e.g., GPS and biometrics), (2) collection of event data (e.g., shots with information regarding shooting coordinates, where it landed, which foot was used, among others), and (3) vision-based collection systems (e.g., tracking data which contains the x,y coordinates of soccer players over time). Vision-based collection systems provide automated and, if correctly implemented, high-precision data. Most data collection methodologies in sports are manual, using very inefficient processes. Since processes are not standardized, much of the data obtained by the clubs cannot be shared to create more robust analytic systems, and data is only relevant for a small range of applications.

Computer Vision can already automate some data acquisition procedures in sports; for example, obtaining tracking data from video. Tracking data is one example where a new category of data in sports leads to fast growth in research, serving as input for physics-based systems \cite{spearman_physics-based_2017}, Machine Learning systems \cite{tuyls_game_2021}, fan engagement tools, or analytic frameworks.

Sports literature has substantial indications that using pose data is the next big step for sports data \cite{tuyls_game_2021}. There are several use cases, such as detecting player orientation \cite{arbues-sanguesa_using_2020}, biomechanical analysis, augmented/virtual reality analytics \cite{rematas_soccer_2018}, and injury prevention, to name a few. 

Therefore, it is vital to understand what Computer Vision tools are available and create value in sports. Automating processes will allow us to extract data quickly and accurately, which will enable many applications that use such data. Furthermore, many data types were previously hard to collect (e.g., gait), which means embracing new ways to collect data in sports can open many future possibilities. 

To further illustrate the use of data acquired through Computer Vision, we present a use case where we aim to predict the speed of a soccer shot based only on pose data. With simple, plug-and-play Machine Learning tools, we reach a 67\% correlation with shot power. This performance indicates a high potential for exploring the insights that Machine Learning models provide, serving as a basis for potentially building recommendation systems or innovative analytics tools. 

In this article, we make the following contributions: 

\begin{itemize}
    \item Section \ref{sec:litrev} presents a review of the literature in Computer Vision. First, it presents the basic principles of Machine Learning and computer vision. Then it describes advanced topics such as Image Recognition, Object Detection, 2D and 3D Pose Estimation.
    \item Section \ref{sec:expsetup} describes a use case of biomechanics analysis, predicting shot speed using only pose data. Then, Section \ref{sec:resultsdiscussion} presents and discusses the results and what they mean in the context of sports sciences.
\end{itemize}

\section{Literature Review} \label{sec:litrev}

Computer Vision is one of the main beneficiaries of the recent Deep Learning boom \cite{lecun_deep_2015}. Many traditional Computer Vision techniques required sophisticated techniques for image pre-processing \cite{aggarwal_human_2011}. Deep Learning removed the emphasis on image pre-processing to leverage the high computational power available nowadays, leading to enormous performance gains. 

\subsection{Machine Learning and Deep Learning}

To better understand the technologies presented, it is essential to understand the goal of Machine Learning. Machine Learning aims to find a function that maps input data (for example, track condition, caloric intake, and weather) into an output (how an athlete will perform in a 100 meters race). Instead of this function being designed by a programmer, the Machine Learning method is responsible to learn this function by itself, using data. 

Different Machine Learning algorithms have different approaches to finding the mapping function. For example, Linear Regression maps use linear relationships, while Decision Trees maps use conditional statements.

The most flexible learning algorithm is Deep Learning (also known as Deep Artificial Neural Networks or simply Neural Networks). Deep Learning finds highly non-linear relationships in data, which is one of the reasons why this algorithm performs much better than any other algorithm in Computer Vision. The other reason is that Deep Learning scales much better than other algorithms when using large amounts of data.

\begin{figure}[h]
    \centering
    \includegraphics[width=\columnwidth]{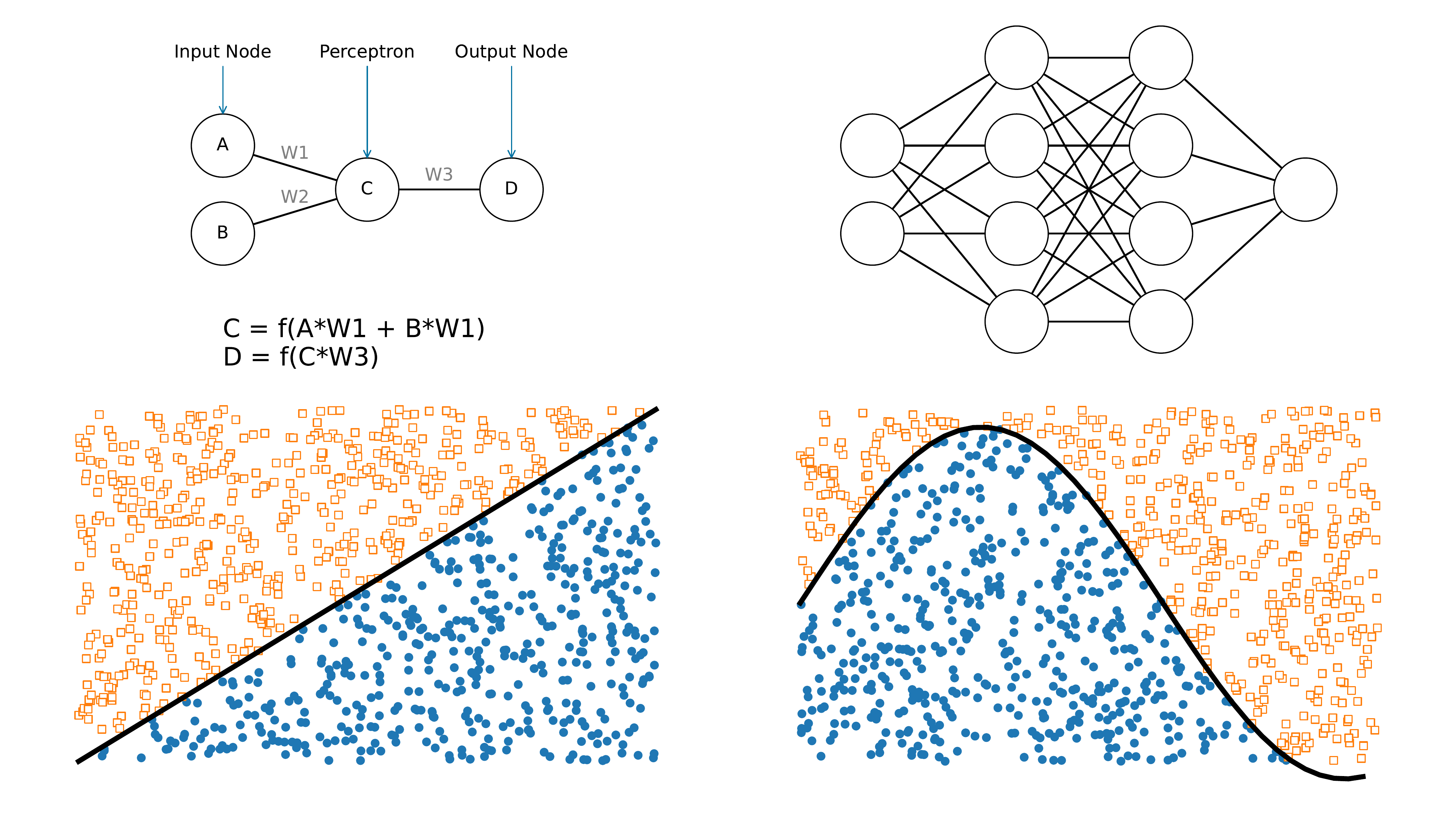}
    \caption{Comparing the Perceptron with a Multi-layer Perceptron. We observe that the Perceptron separates linearly-separable classes, but we require multiple interconnected Perceptrons to separate non-linearly-separable classes.}
    \label{fig:001_neural_network_example}
\end{figure}

The simplest Deep Learning algorithm is the multi-layer Perceptron. Perceptrons are linear classifiers that can distinguish between two linearly separated classes (see Figure \ref{fig:001_neural_network_example}). They are similar to linear regression, multiplying each input by the respective weight. But, in the Perceptron, the output passes through an activation function. The activation function is responsible for adding non-linearities to the prediction.

The activation function alone only models a restricted number of non-linearities. To increase the range of functions that our Perceptron can approximate, we need to combine multiple Perceptrons in a network, organized in interconnected layers. For each link between two interconnected layers, we have a specific weight. The procedure to find the network's weights, usually referred to as training, is an optimization problem and uses algorithms such as Gradient Descent. The optimization of weights is a computationally expensive task, but recent hardware and algorithmic advances enabled training networks with millions of nodes that can perform the Computer Vision tasks we will describe in the following sections.

One Deep Learning architecture is fundamental regarding Computer Vision tasks: Convolutional Neural Networks (CNNs). In Computer Vision, the input given to our models is the raw pixels in an image. Each pixel is a combination of three numbers ranging from 0 to 255. In essence, we are giving our Deep Learning model a matrix whose elements have relationships among themselves. For example, to detect the frontier separating which pixels represent a person and which do not, we need to combine the information from multiple pixels and their respective relative position in an image. This information can substantially simplify the learning procedure.

\begin{figure}
    \centering
    \includegraphics[width=0.35\columnwidth]{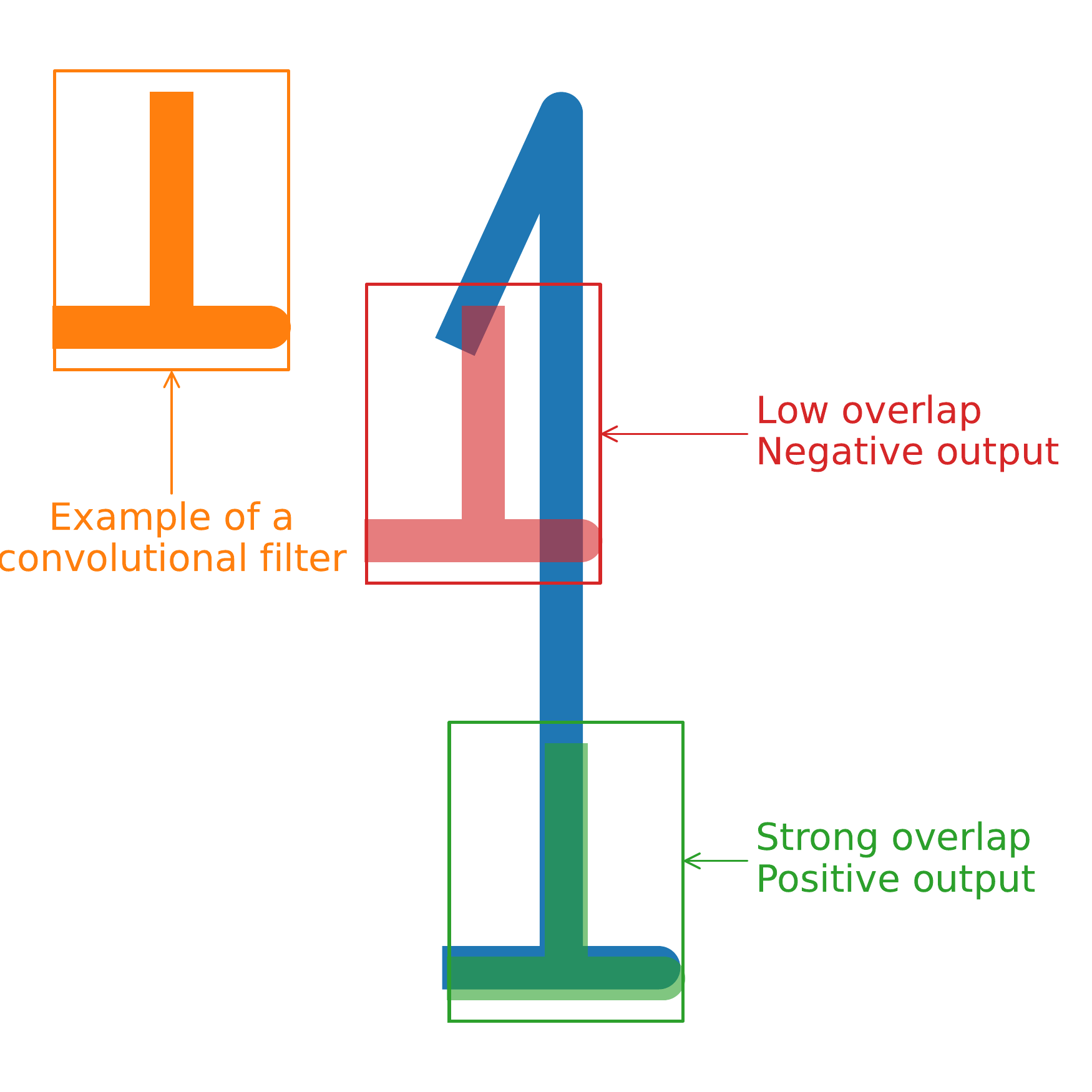}
    \caption{How a Convolutional filter works. We slide the Convolutional filter (in orange) across the image. If the filter matches a feature of the element we want to find, the filter produces a positive output (in green). Otherwise, it produces a negative output (in red). By combining multiple features, we can learn to recognize all hand-written digits.}
    \label{fig:002_convolution_example}
\end{figure}

CNNs solve this problem by applying filters across an image. For example, in Figure \ref{fig:002_convolution_example}, we try to find a feature representing the bottom of the number 1 across an image of a 1. Where the image contains the said feature, the Convolutional Layer of the Neural Network produces a positive output. Otherwise, it gives a negative output. Then, the rest of the Neural Network can use information about which features are detected and where they are located to indicate which number is present in the image. As with the weights of the Deep Learning model, we can also learn the shape for each filter. 

For a deeper explanation of any Deep Learning concepts, see \cite{lecun_deep_2015, aggarwal_neural_2018}.

\subsection{Computer Vision Basics}

\subsubsection{Datasets}    \label{sec:datasets}

As we described before, Computer Vision is nowadays intrinsically linked with Deep Learning. Therefore, there is a high correlation between the efficacy of Computer Vision methods and the amount and quality of data available to train these models. 

Deep Learning learns to map input spaces into the desired outputs. The larger the input space, the larger the training dataset we require. A 1080p image contains more than 6M numbers. Therefore, to build accurate Computer Vision models, we require massive datasets. Table \ref{tab:datasets} lists the most relevant datasets used to build the state-of-the-art models presented in this work. 

\begin{table}[h]
    \caption{Most relevant publicly available datasets for Computer Vision research.}{
    \resizebox{\columnwidth}{!}{
        \begin{tabular}{lccccc} \hline
        \textbf{Dataset} & \textbf{Task} & \textbf{Year} & \textbf{No. Images} & \textbf{No. Classes} & \textbf{Homepage} \\ \hline
        OpenImages \cite{kuznetsova_open_2020} & Object, Location & 2017 & 9M & 19 957 & storage.googleapis.com/openimages/web/index.html \\
        COCO \cite{fleet_microsoft_2014} & Object, Location, Pose & 2015 & 200k & 80 & cocodataset.org \\
        ImageNet \cite{russakovsky_imagenet_2015} & Object, Location & 2014 & 1.4M & 1 000 & image-net.org \\
        SUN \cite{xiao_sun_2010} & Object & 2010 & $\sim$131k & 899 & vision.princeton.edu/projects/2010/SUN \\
        PASCAL VOC \cite{everingham_pascal_2010} & Object & 2007 & $\sim$11.5k & 20 & host.robots.ox.ac.uk/pascal/VOC \\
        MPII HumanPose \cite{andriluka_2d_2014} & Pose & 2014 & 25k & 410 & human-pose.mpi-inf.mpg.de \\
        Leeds Sports Poses \cite{johnson_clustered_2010} & Pose & 2010 & 10k & - & dbcollection.readthedocs.io/en/latest/datasets/leeds\_sports\_pose\_extended.html \\
        3DPW \cite{ferrari_recovering_2018} & Pose (3D) & 2018 & 51k & - & virtualhumans.mpi-inf.mpg.de/3DPW/ \\
        Human3.6M \cite{ionescu_human36m_2014}  & Pose (3D) & 2014 & 3.6M & 17 & vision.imar.ro/human3.6m/description.php \\ \hline
        \end{tabular}
    }}
    \label{tab:datasets}
\end{table}

Although Google's OpenImages is the most extensive dataset, it uses machine-labeled data, which is less reliable. Microsoft's COCO is currently the best and most used public dataset for Object Detection and 2D Pose Estimation. Leeds Sports Poses contains data related to sports, making it a useful dataset for fine-tuning models for sports use cases.

Human3.6M and 3DPW are the go-to datasets for 3D Pose Estimation. While Human3.6M provides more training data with diverse poses and actions, the dataset is captured in a controlled environment (laboratory). Meanwhile, 3DPW provides an in-the-wild dataset, which is very useful for evaluating the models' performance in real-world scenarios. 

PapersWithCode\footnotemark[1] platform presents multiple benchmark rankings for each of the datasets described in Table \ref{tab:datasets}, where the users can find the best performing algorithms and their related literature. While we provide up-to-date information, note that this field is rapidly advancing. Therefore, please refer to this platform to find the best models available now.

\footnotetext[1]{paperswithcode.com} 

\subsubsection{Metrics}

Another relevant part for training and evaluating the model's performance is the evaluation metrics. The datasets presented in Section \ref{sec:datasets} are imbalanced: they provide very few instances with some objects while providing many instances with persons. Learning on imbalanced datasets requires using adequate metrics. For example, we can obtain high accuracy while the models ignore objects with low representation. If the dataset contains 95\% persons and 5\% objects, a model that is 100\% accurate for detecting persons and 0\% accurate when detecting objects, would achieve an accuracy of 95\%, without actually detecting a single object.

\begin{figure}[h]
    \centering
    \includegraphics[width=\columnwidth]{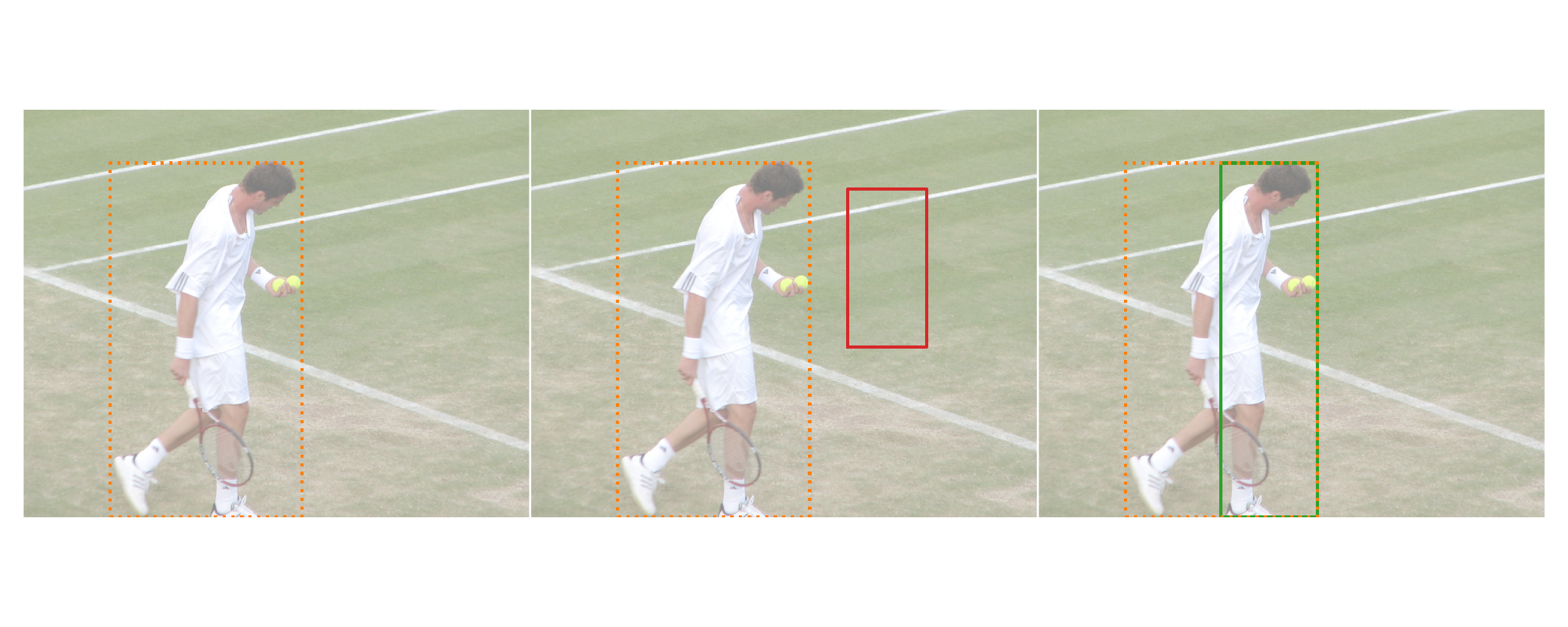}
    \caption{From left to right: (1) The box of the person on the image, (2) a situation where the IoU is 0, since the boxes do not intercept each other, and (3) a situation where IoU is 0.5. Image obtained from the COCO dataset\cite{fleet_microsoft_2014}.}
    \label{fig:003_IOU_example}
\end{figure}

Since Object Detection works by predicting the bounding box around an object (see Figure \ref{fig:003_IOU_example}), we need to first define how we measure a successful prediction. To measure how well a model found the bounding box, we calculate the Intersection over Union (IoU), Equation \ref{eq:iou}. 

Imagine we have two bounding boxes (we will call it box/boxes from now on), one being the ground-truth and the other being the prediction. The Area of Intersection is where both boxes coincide, and the Area of Union is the area covered by at least one of the boxes. If the IoU is above a predefined threshold, we consider the prediction successful. 

\begin{equation} \label{eq:iou}
    Intersection\ over\ Union = \frac{Area\ of\ Intersection}{Area\ of\ Union}
\end{equation}

\begin{figure}
    \centering
    \includegraphics[width=0.8\columnwidth]{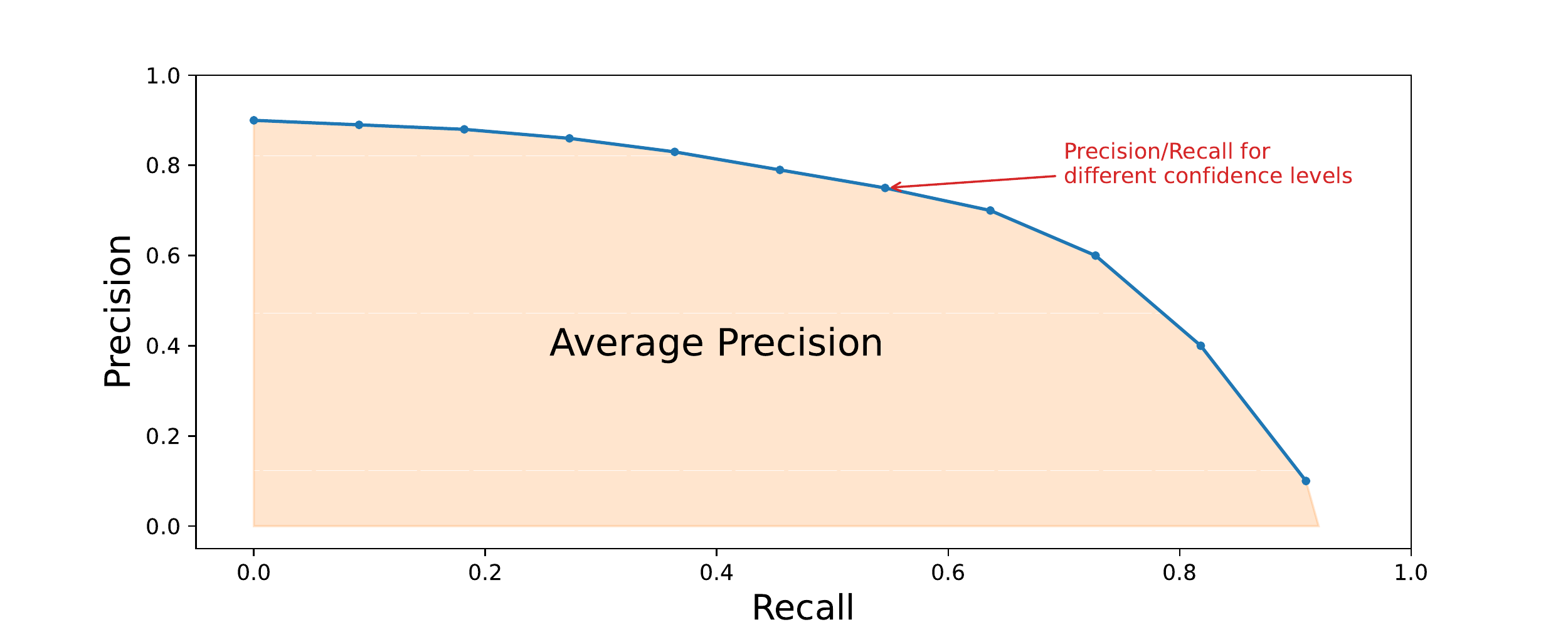}
    \caption{Visualization of the AP metric, estimated by the area below the Precision-Recall curve at different confidence intervals.}
    \label{fig:004_average_precision_example}
\end{figure}

The metric used to avoid the imbalanced data problem is the Average Precision (AP) \cite{su_relationship_2015}, presented in Equation \ref{eq:avgprecision}, where $p(r)$ is the relationship curve between precision and recall. AP corresponds to the area below the Precision-Recall curve (see Figure \ref{fig:004_average_precision_example}), averaged across all classes, at different confidence intervals. 

\begin{equation} \label{eq:avgprecision}
    AP = \int_{0}^{1} p(r) dr
\end{equation}

Since we can arbitrarily define the IoU threshold for considering a successful prediction, the consensus is to calculate the mean of the AP across multiple IoU values, known as the Mean Average Precision (mAP). For example, the COCO dataset \cite{fleet_microsoft_2014} suggests averaging the AP metric for all IoU thresholds between 0.5 and 0.95 with step 0.05. Older datasets, such as PASCAL VOC \cite{everingham_pascal_2010}, suggested using only AP with IoU=0.5. 

For 2D Pose Estimation, we can still use mAP over the body parts boxes. However, our goal in 2D Pose Estimation is not to find the individual boxes that make a person. Therefore, there are better metrics that allow us to verify if we are improving in the right problem, predicting the keypoint location. 

The most used metric is the Percentage of Correct Keypoints (PCK). Like in Object Detection, we need to define how we evaluate if a keypoint is correct. For this, we consider a keypoint correct if it meets a predefined error tolerance. For example, the most used error tolerance is half of the person's head size, referred to as PCKh@0.5 metric. 

In 3D Pose Estimation, using mAP is not possible: both boxes are now 3D, and manually labeling 3D boxes would be a challenging task. PCK metrics are still usable. However, since in 3D Pose Estimation, we must predict every keypoint (and not just the visible), we have access to more informative metrics, like Mean Per Joint Position Error (MPJPE) and Mean Per Joint Angular Error (MPJAE). 

MPJPE is the average distance between each predicted point and actual keypoint across all keypoints, as seen in Equation \ref{eq:mpjpe} \cite{ionescu_human36m_2014}, where N is the number of keypoints predicted, $k_p$ is the predicted keypoint and $k_{gt}$ is the ground truth keypoint. MPJAE follows the same formula but uses the distance between the predicted angle and the actual keypoint angle. 

\begin{equation} \label{eq:mpjpe}
    MPJPE = \frac{1}{N} \sum_{1}^{N}\left \| k_p - k_{gt} \right \|
\end{equation}

With the metrics defined, we can now better understand the models' limitations and consider the measuring error in our analysis. Similar to errors in sensors, models have measurement errors, for which we need to account. 

\subsection{Image Recognition}

In Image Recognition, our goal is to detect which object is in the image. For example, whether an image contains a person or not. In this approach, we do not require any localization of the detected object. Thus, this is a relatively limited approach but is relevant since many Object Detection frameworks leverage the technology behind Image Recognition. 

\subsubsection{Non-Deep Learning Approaches}

CNNs provide a crucial building block for Computer Vision models. With CNNs, we can extract features from the image, which helps the models distinguish between different images. However, CNNs have only become widespread since 2012. Before this, most Computer Vision techniques focused on correctly defining the features from the images so that Machine Learning models could distinguish the different classes, which formed the backbone of Computer Vision models. This section will present the three most widely used methods for feature extraction from images that do not rely on Deep Learning.

\begin{figure}[h]
    \centering
    \includegraphics[width=\columnwidth]{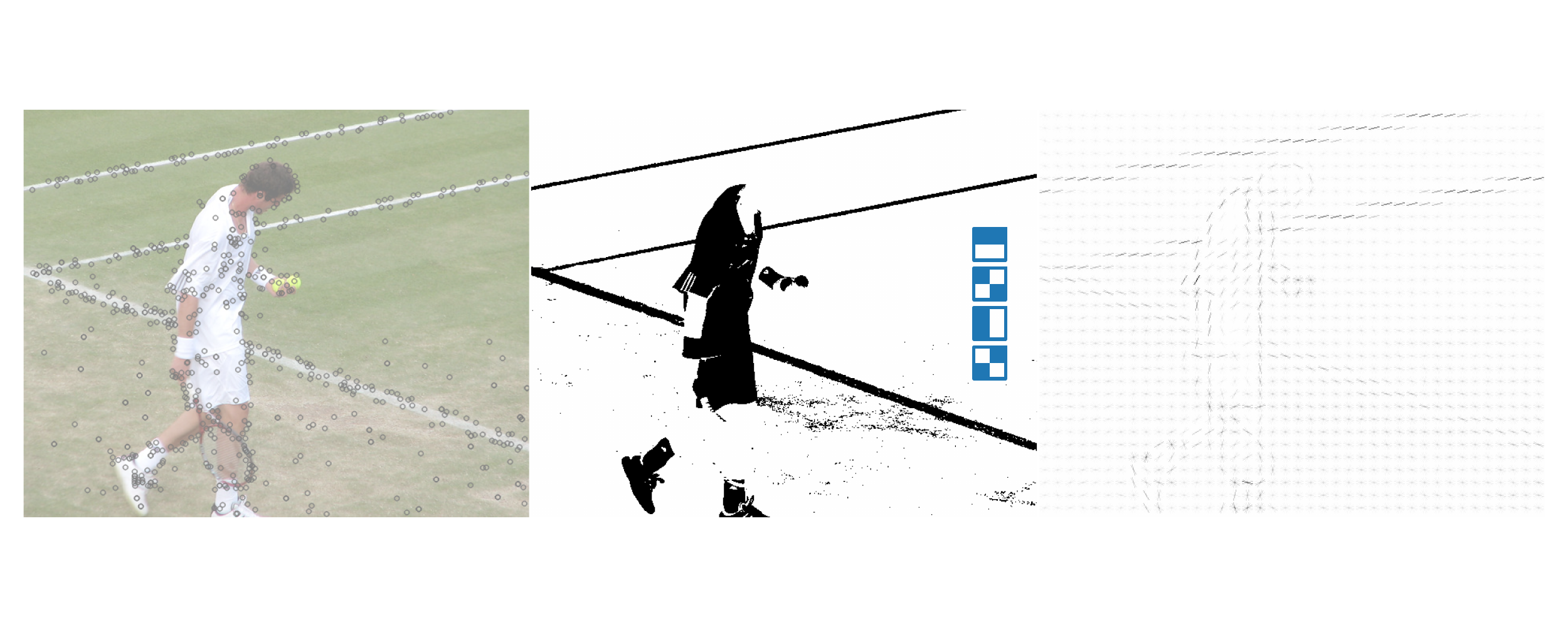}
    \caption{From left to right: (1) Interest points detected by the SIFT method, (2) examples of Haar features (in blue) that can characterize specific features in the binary image, and (3) orientation of the gradients in a image, which are then grouped in one histogram according to the HOG method.}
    \label{fig:005_non_deep_example}
\end{figure}

David Lowe \cite{lowe_object_1999, lowe_distinctive_2004} proposed a feature generation method called Scale-Invariant Feature Transform (SIFT). SIFT detects interest points summarizing the local structures in the image and builds a database with histograms that describe the gradients near the interest points (see Figure \ref{fig:005_non_deep_example}).

This procedure is performed across different image scales, which increases the robustness of image scaling. Also, rotation/translation mechanisms for the gradients introduce robustness to changes of those two components. The method is also robust to illumination changes since we ignore the magnitude of the gradients, using only their orientation.

Viola and Jones \cite{viola_rapid_2001, viola_robust_2004} proposed a facial recognition framework based on generating Haar-like features that aim to capture patterns that are common across a specific type of object. Haar features are rectangular filters used across an image to detect regions where pixel intensities match the layout of the Haar feature. The method is efficient and is still used in facial recognition today but has limited use cases for other Image Recognition tasks. It does not generalize properly to other tasks because it suffers from the influence of rotation of the object and improper lighting.

Dalal and Triggs \cite{dalal_histograms_2005} applied Histogram of Oriented Gradients (HOG) to detect persons in images. HOG creates a grid of points and then calculates the gradient of the pixels in the neighborhood. Then, using only the orientation of the gradients, HOG finds the histogram of orientations that summarizes the whole image. The significant difference to David Lowe's technique \cite{lowe_object_1999, lowe_distinctive_2004} is that HOG tries to summarize the whole image rather than finding specific features in an image. Furthermore, computing HOG features is substantially more efficient than computing SIFT features.

The methods presented are limited in terms of applications nowadays. However, it is interesting to note that they all have the same goal: transform an RGB image into a data structure easily read by a Machine Learning algorithm, which is a procedure that CNNs automate.

\subsubsection{Deep Learning Approaches}

The beginning of CNNs is attributed to Fukushima \cite{fukushima_neocognitron_1980}, who proposed a mechanism allowing Neural Networks to learn features "unaffected by a shift in position", called Neocognitron. The Neocognitron was the first to use Convolutional Layers to construct a feature map by applying filters to the input space.

Later, LeCun et al. \cite{lecun_backpropagation_1989} provided a significant innovation: by applying the backpropagation algorithm, LeCun enabled the Neural Network to automatically learn each filter's weights in the Convolutional layer. Instead of requiring the user to define the filters that create the feature map, the process of defining the filters is also automated. 

It was only much later that CNNs achieved state-of-the-art performance in image classification on the ImageNet dataset with AlexNet \cite{krizhevsky_imagenet_2017}. This model provided a leap in performance over previous state-of-the-art techniques that were dominant at the time (non-Deep Learning approaches), with 42\% error reduction.  

AlexNet used Convolutional Layers to extract features and then classified the image using three extra Convolutional Layers. The most relevant advances of AlexNet were not the network architecture but rather the computational tricks that allowed training such a large model. First, it used ReLU as an activation function, which enables much faster training than other options. Second, the architecture enabled training the model in parallel using two graphics processing units instead of one, enabling faster training times and larger models.

In the following years, the models became increasingly complex but faced one problem: performance degraded when the number of layers increased past a certain threshold. This degradation went against the consensus that a network with N1+N2 layers should match the performance of a network with N1 layers simply by learning the identity function in the remaining N2 layers. However, as networks grew deeper, learning even the identity function became very difficult.  

He et al. \cite{he_deep_2016} solves this problem by introducing skip connections that give an alternative path, guaranteeing that a deeper network can consistently perform at least as well as a smaller network. The called ResNets serve as a backbone for many methods on both Object Detection and Pose Estimation. Even by themselves, ResNets produced state-of-the-art performance in Image Recognition at the time of their publishing. 

The approaches that we described in this section until now fail to address the localization aspect. They focus on Image Recognition rather than Object Detection. The state-of-the-art techniques that we will present next all address both the classification and localization problem.

\subsection{Object Detection}

Object Detection adds a step to Image Recognition: locating the coordinates of an object in the image. Szegedy et al. \cite{szegedy_deep_2013} framed Object Detection as the problem of predicting object's boxes by regression but achieved limited success with this approach. The most successful approach is to find candidate regions/boxes for objects and then treat the region as a single image. 

Girshick et al. \cite{girshick_rich_2014} introduced this latter approach with a method called Region-based CNNs (R-CNN). The method achieved an AlexNet-like leap in performance, increasing ~53\% over the previous best approaches. R-CNN follows three steps: (1) recommending candidate boxes/regions, using Selective Search \cite{uijlings_selective_2013}, (2) computing features using CNNs, and (3) classifying each region using Support Vector Machines (SVM). Selective Search generates many candidate regions and then greedily combines similar regions into larger ones. 

Iteratively, the R-CNN architecture received many improvements. Some improvements are related to improving the speed, like Fast R-CNN \cite{girshick_fast_2015} which computes the feature map for the whole image instead of computing for each candidate region, accelerating training and testing. Then, the Faster R-CNN \cite{ren_faster_2017} introduced a Region Proposal Network that accelerates the process of providing region proposals. Instead of using Selective Search, Ren et al. \cite{ren_faster_2017} uses a Deep Learning model that predicts the region location. 
Other proposed solutions focused on improving performance, like Retina-Net \cite{lin_focal_2017} and Libra R-CNN \cite{pang_libra_2019} which change training to focus on hard-to-learn objects instead of focusing uniformly across objects, which addresses the imbalance issues in the data set. 

\begin{figure}[h]
    \centering
    \includegraphics[width=\columnwidth]{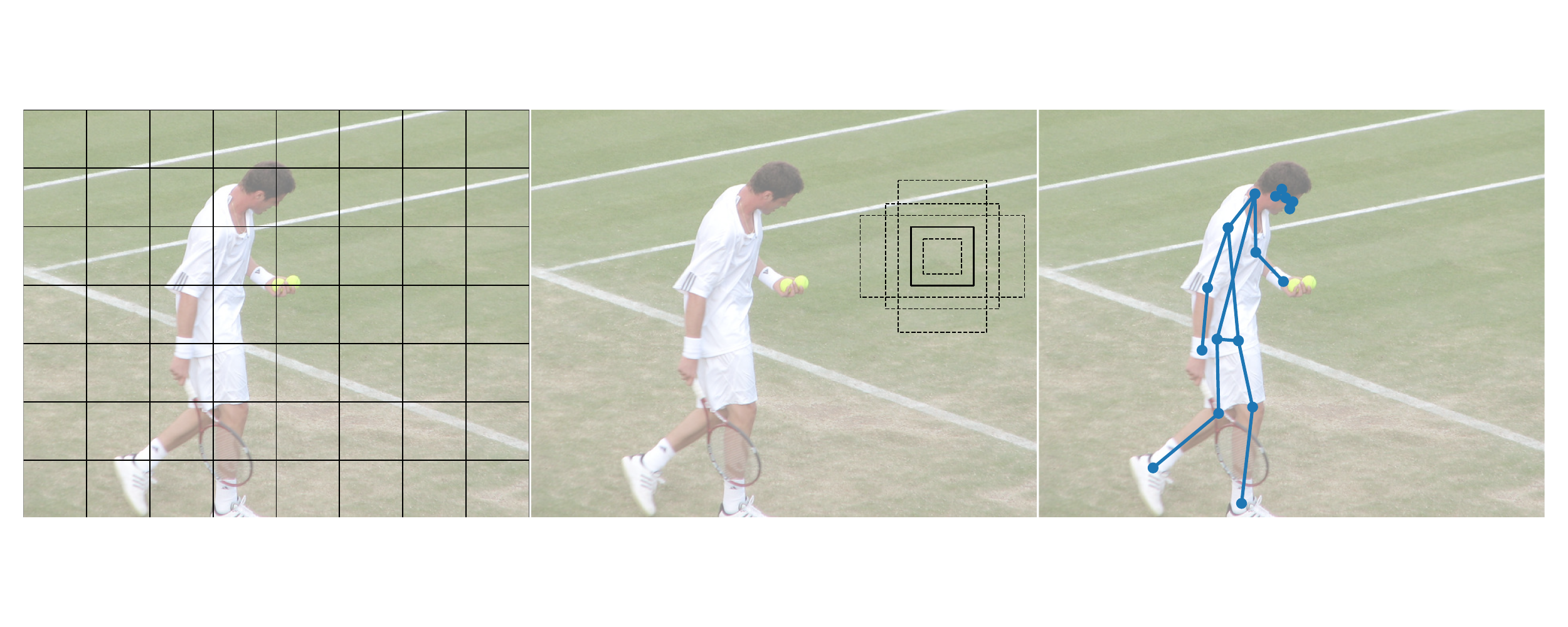}
    \caption{From left to right: (1) Grid-style boxes used by the original YOLO algorithm. (2) An example of the default box styles used by RefineDet. (3) Example of the keypoints used by the CenterNet framework.}
    \label{fig:006_object_concepts_example}
\end{figure}

The major drawback of R-CNN-based models is that they require different frameworks to detect the regions and classify them. Redmon et al. \cite{redmon_you_2016} proposed using a single model to detect and classify regions. The method proposed, You Only Look Once (YOLO), draws a grid of cells covering the whole image without intersecting each other (see Figure \ref{fig:006_object_concepts_example}). Then, each grid cell generates multiple candidate boxes and a prediction of whether the box contains an object or not. Then, it combines the generated candidate boxes with the respective cell's prediction of which object is in the cell to output the final prediction.

YOLO has a large development community, with numerous improvements made on the same proposed architecture. YOLOv2 to YOLOv5 \cite{redmon_yolo9000_2017,redmon_yolov3_2018,bochkovskiy_yolov4_2020}, YOLOR \cite{wang_you_2021}, PP-YOLO v1 and v2 \cite{long_pp-yolo_2020,huang_pp-yolov2_2021} provided improvements to the model accuracy and prediction speed. The Single Shot Detector (SSD) \cite{leibe_ssd_2016} follows a similar approach from YOLO. The most significant distinction between SSD and YOLO is that SSD uses predefined starting boxes, which the model adapts using offsets and scale factors during the prediction process. 

Zhang et al. \cite{zhang_single-shot_2018} proposed combining both the best of R-CNN multi-phased approaches (that lead to better accuracy) and single network proposals (that are faster to train and predict). For this, he proposed a box-based single-shot network (similar to SSD), RefineDet, which contains a box refinement module. This box refinement module assumes default box styles (see Figure \ref{fig:006_object_concepts_example}). The approach removes irrelevant boxes and adjusts the location of the remaining.

While the single-network approaches are much quicker in prediction, in some use cases, we require even faster prediction times. For this reason, there are multiple proposed mobile models based on single-network approaches, which have the goal of scaling down networks to make faster predictions with limited resources. 

EfficientNets \cite{tan_efficientnet_2019} defines a scaling method that allows efficient scaling of both network size and image resolution efficiently. EfficientDet \cite{tan_efficientdet_2020} further optimized EfficientNets with a method to use the same features across different scales. The YOLO-related literature also contains many proposals for efficient models called TinyYOLO. 

Zhou et al. \cite{zhou_objects_2019} proposed CenterNet, which uses an approach that focuses on modeling the object as the center point of its box. CenterNet predicts the center of the box and keypoints of the object, for example, joint locations on human bodies (see Figure \ref{fig:006_object_concepts_example}). The keypoints are crucial for this method since they expand the box from the predicted center. One of the main advantages of this method is that it also yields the object keypoints, which increases the range of possible use cases.

To summarize, we have two different solutions: (1) R-CNN architectures that are very good but slow, and (2) single-network architectures that are fast but have lower prediction power. 

The most recent state-of-the-art has shifted from CNNs towards Transformers. Transformers are Attention-based Deep Learning architectures. While CNNs are excellent for discovering local features, Transformers are excellent for finding global features. For this, it uses Attention, a technique that enhances important parts of the image for prediction while reducing the importance of the rest. 

While Attention was proposed by Vaswani et al. \cite{vaswani_attention_2017} in 2017, it was difficult to use it in Computer Vision. The reason is that since these techniques calculate the importance of each input pixel, it would be costly to perform in images that have millions of pixels. To make matters even worse, the number of operations does not increase linearly with the number of pixels but quadratically. Dosovitskiy et al. \cite{dosovitskiy_image_2021} proposed splitting images into fixed-size patches, which made Transformers efficient, requiring less computational resources than CNNs, even when controlling for the lower number of parameters in models. 

Transformer-based models perform better when trained in super-large data sets because they can continue improving past the point where CNNs hit their plateau. On the other hand, CNNs are more data-efficient because they can reach better performance levels when working with limited (but still large) datasets \cite{dosovitskiy_image_2021}. 

\begin{figure}[h]
    \centering
    \includegraphics[width=\columnwidth]{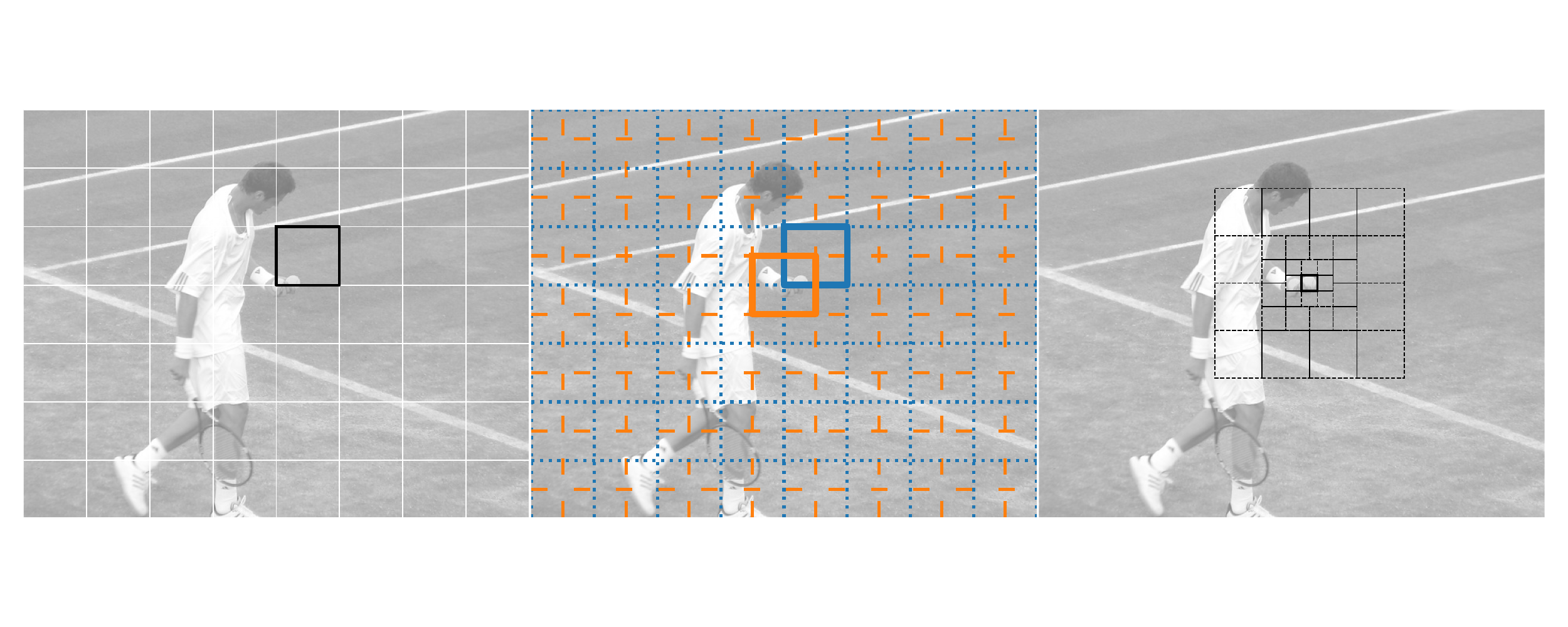}
    \caption{From left to right: (1) The original size of a patches allowed very low resolutions, lacking precision when detecting the objects. For example, the box corresponding to the tennis ball would be the dark box. (2) Using a shifting window increases the resolution of the method by shifting the patches over iterations. By combining two patches, we can increase the resolution of the method substantially. (3) To reduce the cost of using Attention, we can use small patches for the close neighborhood and larger patches to distant points in the image.}
    \label{fig:007_attention_concepts_example}
\end{figure}

Many proposals aim to solve the problems using Transformer-based architectures. Liu et al. \cite{liu_swin_2021} proposed using shifting windows and patch merging to artificially increase the resolution of the technique, which was previously limited to the size of the patches (see Figure \ref{fig:007_attention_concepts_example}). This change allowed a higher granularity in the predictions, which we require for Object Detection. Patch merging also reduced the computational complexity from quadratic to linear by computing Attention within the respective patch. 

Efficient patch usage is one of the most focused areas. Yang et al. \cite{yang_focal_2021} proposes using dynamic patch sizes according to the distance to the patch in which we calculate the Attention. This way, neighbor patches are smaller than further away patches, decreasing the number of patches to process (see Figure \ref{fig:007_attention_concepts_example}), and accelerating the procedure while maintaining performance.

Transformers also allowed some simplified architectures to perform very well in Object Detection. Xu et al. \cite{xu_end--end_2021} proposes treating Object Detection as a prediction problem by introducing a "no object" class which allows the network to predict a fixed number of objects for each image. 

While some studies have discarded CNNs entirely in favor of Transformers, some works focused on using Attention to enhance current CNNs architectures. For example, Dai et al. \cite{dai_dynamic_2021}, which explores the introduction of Attention in the later stages of the models (the prediction phase), allowing improvements using the existing models' backbones (i.e., part of a Deep Learning model that is responsible for generating the features). 

Attention-based methods have seen significant advances in the last year, and their performance already surpasses methods using only CNNs. The most effective methods combine both Attention and CNNs to produce the best results in terms of accuracy.

One last notable work is from Liang et al. \cite{liang_cbnetv2_2021}, which proposes a framework for combining multiple model backbones. This ensembling approach increased the accuracy of the models over using only one isolated model, even when combining only two models. 

Pretrained versions of most models presented in this section are available in libraries such as the TensorFlow Hub\footnotemark[2], DarkNet\footnotemark[3], or published by their authors in public repositories. These versions are plug-and-play versions of the models that are easy to use across multiple applications. Furthermore, benchmarks of most of the approaches are available on the PaperWithCode platform. 

\footnotetext[2]{tfhub.dev/tensorflow/collections/object\_detection} 

\footnotetext[3]{pjreddie.com/darknet} 

\subsection{Pose Estimation}

\begin{figure}[h]
    \centering
    \includegraphics[width=\columnwidth]{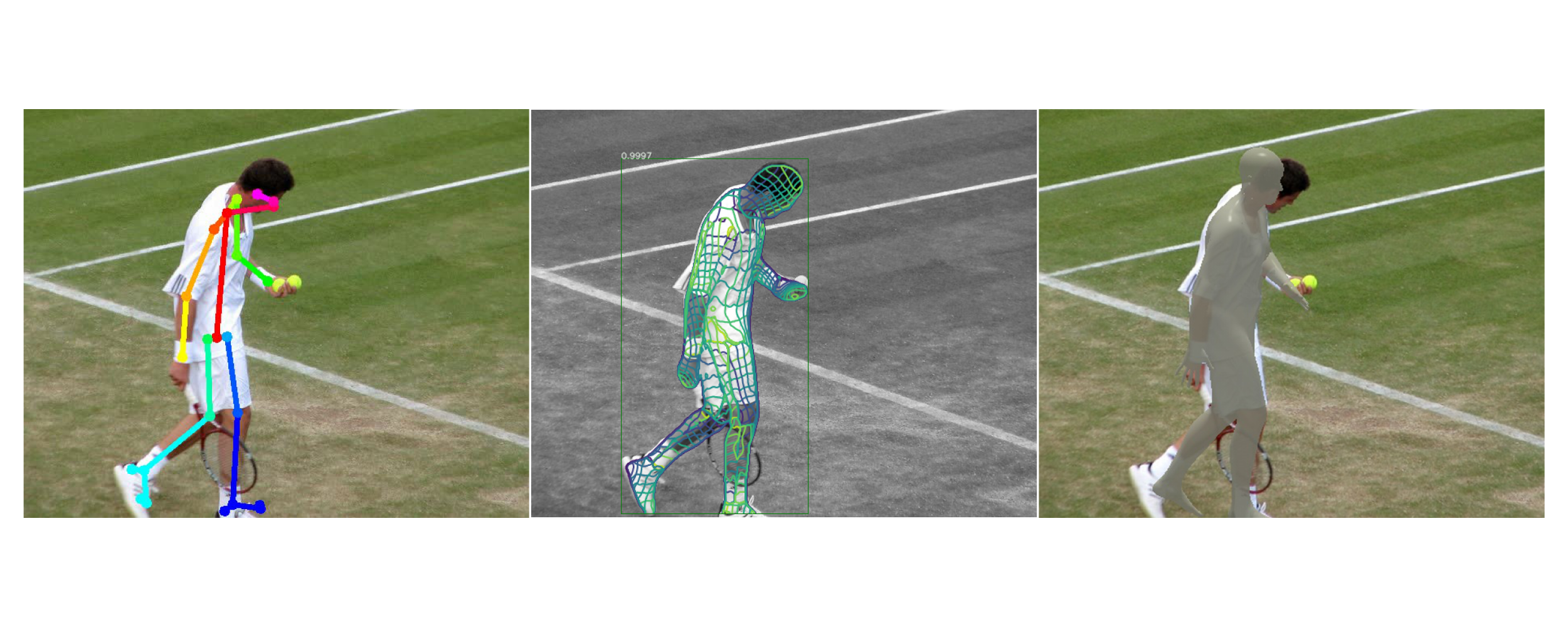}
    \caption{From left to right: (1) Example of the keypoints in the COCO dataset, as obtained by OpenPose. (2) Example of the output of a DensePose model. (3) An example of a 3D mesh as described by SMPL.}
    \label{fig:008_keypoint_example}
\end{figure}

Pose (or Gait) Estimation (or Extraction) methods aim to detect the key body parts/joints that define the human body in an image. As a result, these methods give us the coordinates of specific human body parts in an image. The methods in this category require images with the keypoints labeled for training, which are significantly harder to label when compared to the Object Detection problems. We can see an example of the keypoints in the COCO dataset in Figure \ref{fig:008_keypoint_example}.

In this section, we cover Pose Estimation methods using Deep Learning. Since the field is evolving towards using multi-person pose estimators (in opposition to single-person pose estimators), we do not use this to categorize the approaches. Instead, we prefer to categorize them according to the two main approaches: (1) top-down, where we detect the box of the person and then detect the keypoints within the box, and (2) bottom-up, where we detect body parts in an image and then aggregate them to form a person.  

Both approaches leverage similar methods to the ones described in the Object Detection section to detect the box of a person or body parts. Furthermore, CNNs are still the basis of the network's layers for feature generation, with some improvements in specific use cases \cite{tompson_joint_2014, wei_convolutional_2016,leibe_stacked_2016,sun_deep_2019}. 

Top-down approaches allow for easy use of single-person pose estimators and Object Detection methods that detect the boxes of persons in an image. This method has a clear drawback: when two people overlap in the image, it struggles to detect which body parts belong to which person.  

Regarding this approach, we highlight the following contributions: Papandreou et al. \cite{papandreou_towards_2017} presented an approach that uses the Faster R-CNN for box detection and then uses a ResNet backbone to build the keypoint estimation model inside the box. Chen et al. \cite{chen_cascaded_2018} combined two networks: one to predict keypoints with high confidence, and another network specializes in detecting harder keypoints, such as occlusions and out-of-bounds keypoints. Carreira et al. \cite{carreira_human_2016} introduced a model that, instead of predicting the keypoints, outputs a suggestion of correction for the current keypoints. Iteratively, this work corrects an initial standard pose towards the actual pose by using the model's feedback, with the number of iterations being user-defined. 

Bottom-up approaches focus on correctly identifying body parts in an image and then grouping them. The procedure of identifying a body part in an image uses the same strategies as in Object Detection. The main challenge of the bottom-up approaches is to correctly group body parts with varying degrees of difficulty. For example, identifying how to group a single body is easy, but for multi-person detection, the problem is substantially harder, especially if they overlap in the image. 

DeepCut \cite{pishchulin_deepcut_2016}, and DeeperCut \cite{leibe_deepercut_2016} use pairwise terms to understand which body parts are likely to belong to the same person. OpenPose \cite{cao_realtime_2017} uses part affinity fields to merge different keypoints. Part affinity fields are flows that indicate the orientation of a body part which allows for a more accurate joining procedure. OpenPose \cite{cao_openpose_2021} also extends the number of keypoints detected; while other approaches only detect the most prominent joints in the human body, this approach also allows detecting hand and facial keypoints. 

The approaches presented above use similar keypoint structures, usually following the labels in the COCO dataset. However, DensePose \cite{guler_densepose_2018} proposed a new approach for Pose Estimation: it uses an annotation pipeline that allows for image annotation at the surface level. This approach will enable algorithms to estimate poses in a different, enriched format (see Figure \ref{fig:008_keypoint_example}). 

Overall, the bottom-up approaches are more successful in 2D Pose Estimation. When both Object Detection and Pose Estimation are required, some methods provide both solutions in one model, leading to quicker prediction time than running each method individually, as the CenterNet approach described in the last section.

\subsection{3D Pose Estimation}

Although 2D Pose Estimation is helpful for many use cases, it is a limitation in others. For example, 2D Pose Estimation may be used in gyms to track posture in specific movements but with the restriction of analyzing only one point of view. We can use it to measure the depth of a squat, or whether the quadriceps or the glutes activate first by checking the Pose Estimation of a side view, but we will lose on detecting potential side imbalances that we would find on the front or back views. Furthermore, if we try to film from an angle that aims to capture both, we will lose accuracy compared to an adequate filming angle. 

3D Pose Estimation can fix these issues by obtaining 3D data from a single view. With 3D Pose Estimation, we can reduce the loss in accuracy from using a filming angle that aims to capture all details from the movement. Adding this third dimension introduces much more information on the data obtained, resulting in better insights from the analysis. 

Currently, 3D Pose Estimation methods focus on estimating the 3D meshes of the people present in the image. Bogo et al. \cite{leibe_keep_2016} introduced a methodology that uses SMPL \cite{loper_smpl_2015}, a model learned from data that represents the human body shape with pose dependencies. Bogo et al. inferred 3D coordinates from 2D coordinates (extracted using DeepCut) by fitting the SMPL model to the 2D data. The author also discusses that 2D coordinates contain much of the information required to extract 3D poses, which might indicate that, for some applications, acquiring 2D data is a sufficient and lesser complex task. We can see an example of a SMPL mesh in Figure \ref{fig:008_keypoint_example}.

To the best of our knowledge, nearly every relevant method published in the literature relies on mesh prediction, specifically, the meshes described in SMPL \cite{loper_smpl_2015}. SMPL meshes contain information about 3D pose keypoints and body shape, allowing us to detect differences in height and body shape between two persons. 

Sun et al. \cite{sun_monocular_2021} predicts if the pixel belongs to the person's center and predicts the mesh for a person centered in the pixel throughout the whole image. Then it searches for clusters in the predictions, and if a cluster is found, the center of the cluster suggests the predicted mesh.

One of the significant known errors are the biases introduced by data in the models. Guan et al. \cite{guan_out--domain_2021} fixed some of these issues, namely the ones related to camera parameters, backgrounds, and occlusions, and achieves top performance in the 3DPW benchmark, which indicates how important it is to fix these issues. 

While state-of-the-art 2D Pose Estimation did not use Transformers, some 3D Pose Estimators already take advantage of this method. Kocabas et al. \cite{kocabas_pare_2021} increases robustness to partial occlusion using Attention. For each occlusion, the network finds which detected body parts should use to infer the occluded part. For example, if the elbow is occluded, it checks the orientation of the shoulder to estimate where the elbow is. 

Hu et al. \cite{hu_conditional_2021} is one of the few approaches that focus on predicting only the keypoints in the form of a directed graph. This specialization allowed the algorithm to perform very well, being the best performer in the Human3.6M benchmark in single view at the time of writing. 

There are several proposals for 3D Pose Estimation from video. By using multiple video frames as input, we can give temporal information about the pose to the models, thus enhancing the performance \cite{pfister_flowing_2015,girdhar_detect-and-track_2018,kocabas_vibe_2020,li_exploiting_2021}. As a trade-off, these algorithms demand more resources since their inputs are larger. 

Another possibility for enhancing performance is to use multi-view models. This approach adds computational complexity and hardware complexity since it requires at least two points of view. Iskakov et al. \cite{iskakov_learnable_2019} combines 2D keypoints from multiple views into a single 3D representation using algebraic triangulation. Reddy et al. \cite{reddy_tessetrack_2021} combines both the temporal and multi-view approaches, while He et al. \cite{he_epipolar_2020} combines multi-view approaches and Transformers, which can perform the Attention operation across multiple frames. 

\subsection{Other Computer Vision Methods}

At last, we will describe some important methods that can complement some of the methods described above. 

The first is transfer learning \cite{zhuang_comprehensive_2021}. Transfer learning is a methodology to refine models, like the ones we described in previous sections, to perform a different task. For example, many of the proposed algorithms work with on-the-wild images. One can further train the described models on a specific data set related to sports if the goal is to use these methods in sports, enhancing their performance for the task.

The second method is image super-resolution \cite{wang_deep_2021}. Image super-resolution aims to increase the resolution of an image by running it through a model that upscales said image. This method can be useful to correct some deficiencies in the capturing procedure, such as having to film from a high distance or sacrificing image quality for a higher frame rate.

\section{Experimental Setup} \label{sec:expsetup}

The main goal of our experiments is to present a methodology that can predict the speed of a shot using only the pose variation of a player. Suppose we obtain a high correlation between the predicted speed of the shot and the actual speed of the shot. This will enable us to create a system that analyzes the shots from multiple players and learns which technique is the best to kick the ball with high speed, depending on player's physical and technical traits. This methodology can be extended to many other use cases. 

With this system, we can provide feedback to players about their technical movement, providing recommendations that consider an extensive history of the technical movement. The recommendations can be, for example, related to certain timings in the movement, position of indirect body parts, or simply detecting the best run-up strategy for the player. 

In this work, we collect 145 soccer shots from a single person. The shots are performed 6 meters from the goal and captured using 90 frames per second camera. To extract data from the video, we use two algorithms: (1) CenterNet HourGlass104 1024x1024 with keypoints\footnotemark[4] and (2) ROMP\footnotemark[5]. Both models provide state-of-the-art performance for the tasks and are publicly available. Each shot takes in average 650 and 43 seconds to produce results in (1) and (2) respectively. Note that (2) uses GPU while (1) is limited to CPU, which explains the difference in prediction time.

Code and processed data are available at GitHub\footnotemark[6]. 

\footnotetext[4]{tfhub.dev/tensorflow/centernet/hourglass\_1024x1024\_kpts/1} 
\footnotetext[5]{github.com/Arthur151/ROMP} 
\footnotetext[6]{github.com/nvsclub/ShotSpeedEstimation}

\subsection{Calculating the Ground Truth}

\begin{figure}[h]
    \centering
    \includegraphics[width=\columnwidth]{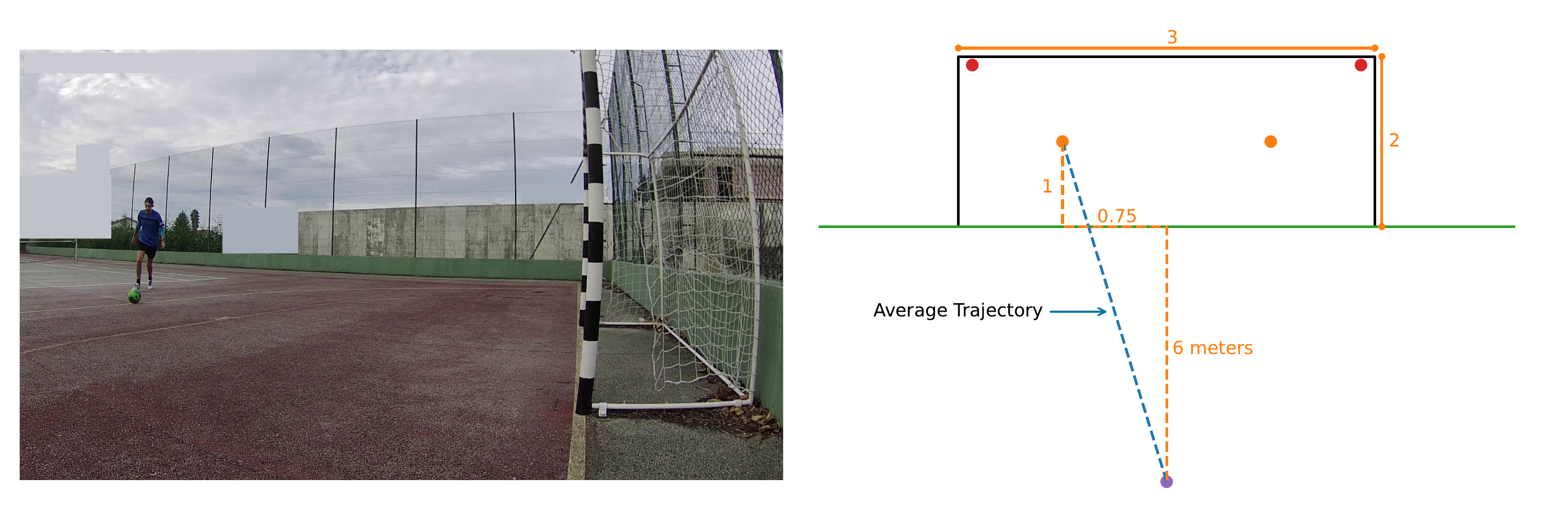}
    \caption{A snapshot of the captured video and the capture procedure.}
    \label{fig:010_experimental_setup}
\end{figure}

To calculate the ground truth (the ball speed), we make two assumptions: (1) the ball hits the center of the left/right side of the goal (see Figure \ref{fig:010_experimental_setup}), and (2) the time the ball takes to travel is calculated between the player's first contact and right before the ball gets occluded by the goalpost.

In the first assumption, we can calculate the maximum error for the ground truth, the difference in distance between the largest trajectory (which we will assume is the corners of the goal) and the average trajectory (the center of the goal). The goal's size is 3 meters in width, 2 meters in height. This assumption introduces an error of +/- 6\% in our measurements. More importantly, the error follows a uniform distribution, meaning that the average error is 0\%. To reduce this error component one could increase the distance to the goal.

We made the second assumption to automate the calculation of the time between the moment the player hits the ball and the moment the ball hits the goal. We use the fact that the ball is occluded when going between the posts to determine our endpoint. Furthermore, we can detect partial occlusions of the ball, ensuring that we can detect the point when it reaches the goal accurately. 

\begin{figure}[h]
    \centering
    \includegraphics[width=\columnwidth]{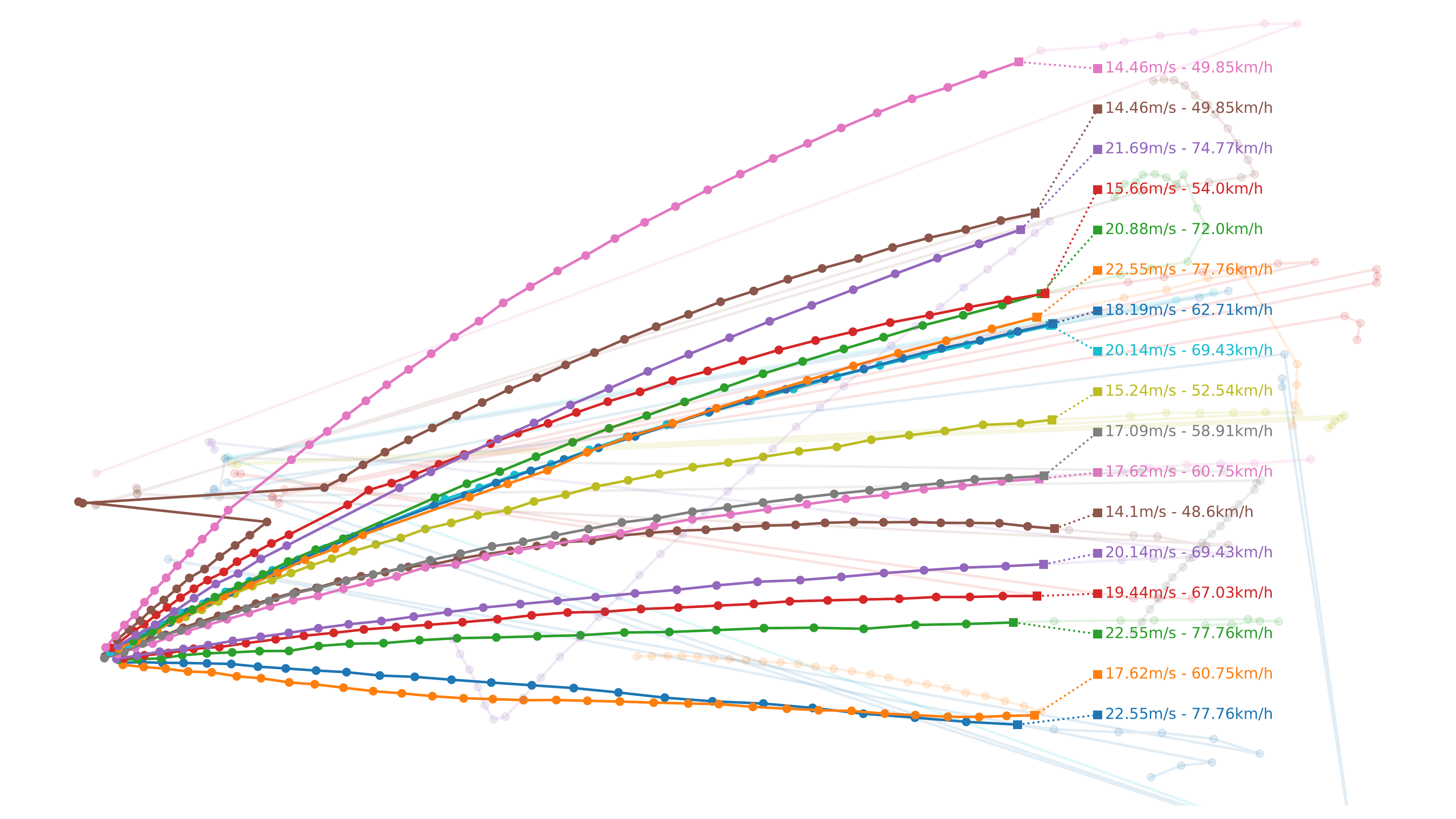}
    \caption{Visualizing a sample of the ball's trajectories as captured by the cameras, along with the calculated speed of the ball.}
    \label{fig:011_ball_trajectories}
\end{figure}

To calculate the time between the shot and the goal, we count the frames up to the point where the ball does not register (occluded by the goalposts) or the ball area is substantially reduced (partially occluded by the goalposts) and divide the count by the frame rate (90fps). For the average speed of the ball, we divide the average distance by the calculated time. To reduce the error caused by this component, we would need to increase the framerate of the camera. The results of this procedure are presented in Figure \ref{fig:011_ball_trajectories}.

\subsection{Processing the Pose Data}

The ROMP model provides many data points that are useful for multiple applications. In our case, we use the SMPL 24-joint data. This format provides information about the keypoints required. 

Our first step was to transform the data that contained absolute positions to angular speeds. According to most of the literature, angular motions tend to be better predictors of good biomechanical performance. Since absolute positions are noisy, we calculate the angular speed from the moving average of the absolute position with a window of 5 frames, which smooths our data points. 

\begin{figure}[h]
    \centering
    \includegraphics[width=\columnwidth]{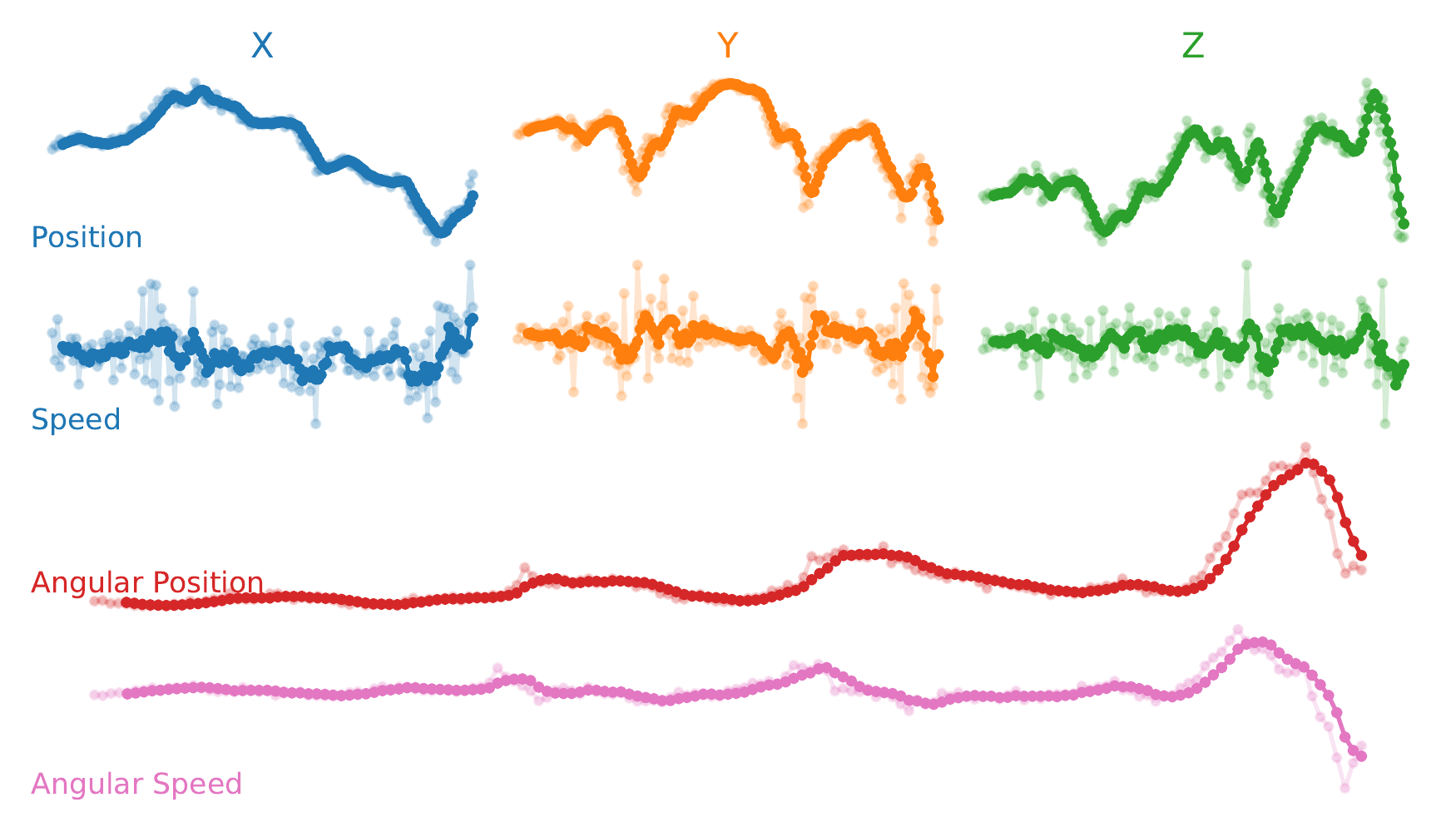}
    \caption{Visualizing the data obtained from the 3D Pose Estimation. In this case, it was obtained from tracking the right knee in a shot at 90 fps. On the top, the absolute position and speed coordinates, which are light in the raw format and darker after smoothing the time series. In the bottom, the angular position and speed calculated from the absolute position and speed time series.}
    \label{fig:012_data_processing}
\end{figure}

However, the translation from absolute positions to angular velocities introduces quite a lot of noise. Therefore, we smooth out the angular time series again using the moving average with a window of 5. Figure \ref{fig:012_data_processing} illustrates the whole process. 

There is one last problem with our dataset. Machine Learning algorithms require inputs of a fixed size. In our case, we could use the time series as input, but time series from different shots have different lengths. Our solution is to extract features from the time series. There are many approaches to extracting features from time series. We will use an automated approach to avoid getting too technical. For this, we use the TSFresh \cite{christ_time_2018}, which transforms time series in hundreds of features. Using the filter function from the same library, which filters the most relevant features for our problem, we obtain 858 features for each shot. 

To compare the difference in results obtained between 2D and 3D pose data, we run the whole processing pipeline on the 2D Pose Estimation data obtained from the CenterNet model, obtaining only 273 relevant features.

\subsection{Predicting the Speed of the Ball}

To create the prediction model for ball speed, we test 4 algorithms: (1) Linear Regression, (2) K-Nearest Neighbors, (3) Gradient Boosting, and (4) Random Forest. We opt to use these algorithms since they provide state-of-the-art results for many other problems, are easy to use, and require little parameter tuning to produce good results. We opt not to use Deep Learning due to the low ratio of instances per feature (1 instance per 5 features), which leads to poor performance of the algorithm. 

To guarantee that the models do not simply memorize data, we use K-Fold Cross Validation, which divides data into train and test sets. By training the model in the train set and hiding the test set data, we ensure that the models try to predict instances that they have not seen before. Both processes using 2D or 3D pose data follow the same procedure.

\section{Results and Discussion} \label{sec:resultsdiscussion}
\begin{figure}[h]
    \centering
    \includegraphics[width=\columnwidth]{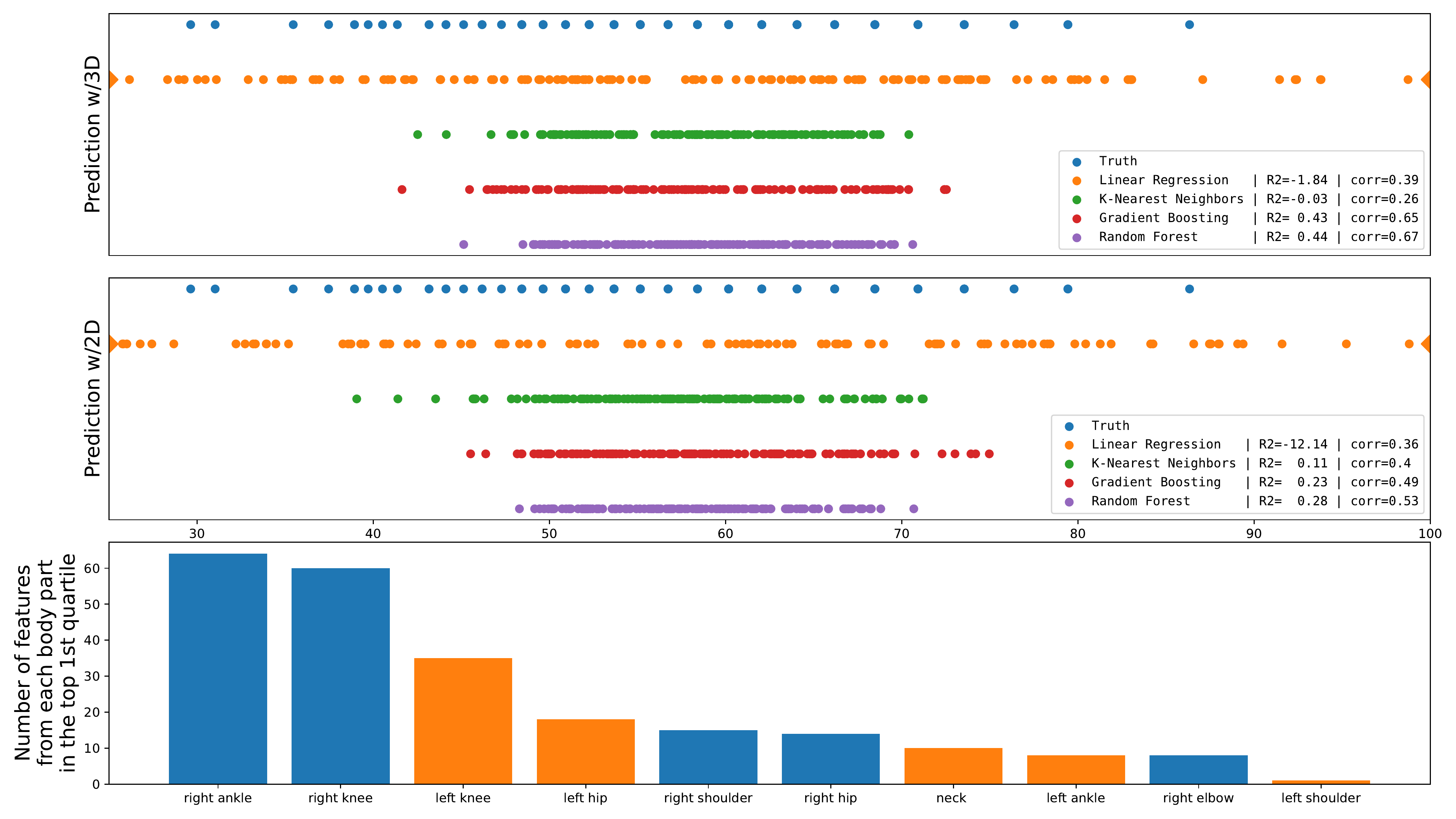}
    \caption{Performance of the models using 3D and 2D pose data, with the Random Forest algorithm achieving 67\% correlation in 3D and just 53\% in 2D. On the bottom, we present the body parts where the Random Forest algorithm focused to obtain the best performance in the 3D problem.}
    \label{fig:013_results}
\end{figure}

Figure \ref{fig:013_results} presents the performance of the models built across multiple dimensions.

By observing Figure \ref{fig:013_results}, we see that the Random Forest model has the best performance between the algorithms tested. Although the r-squared is relatively modest (0.44), it achieves a correlation of 67\% with our target variable.  

Obtaining 67\% correlation is very good, especially considering the low number of shots available, the fact that we did not perform any hyperparameter tuning, and we are using automatic feature engineering. We believe that by addressing these issues, we will substantially improve the quality of the prediction. 

This experiment's main goal is to show that there are many possibilities for using the combination of Computer Vision to acquire data and Machine Learning to model. As these fields mature, it is very important to start working on problems that were previously very hard to solve but are now becoming accessible.  

A high correlation model has tremendous applications, such as building a player recommendation system tailored to a specific technical movement. Furthermore, for elite players, the "save point" created when gathering this data can help them recover performance in specific technical movements in injury recovery or lack of shape scenarios. Injury prevention can also see advantages by using these methods since they can detect slight differences in player technique, indicating a higher probability of injury. 
With only 145 shots, we achieved a 67\% correlation, which is impressive if we consider that we can scale the methodology to capture thousands of shots from multiple individuals. This tremendous amount of data will increase the type of insights we gather from these kinds of experiments. 

Even though we believe this research path is ready for the next step, we acknowledge that there are some limitations. For example, the real-time performance of these methods is not up to the required standards, with errors that make real-time applications unadvised. Another limitation is that we still require a relatively controlled setup. Since many sports, namely collective sports, require on-the-wild detection, we need to ensure we acquire data with the highest quality possible, like correct camera angles and position to avoid occlusions. Having multiple people in the camera range also increases the difficulty of filtering the data from the correct athlete since most methods detect all people in an image.

Another thing to note is that models that use 3D data far outperform the models that use 2D data. Even when considering that 3D data is less precise than 2D data, the extra dimensionality improves the results substantially. Therefore, if the goal is to obtain the best results possible, we should always be using 3D data.

The advance in using this kind of data in biomechanics research can also incentive creating a sports-related data set. A sports-related 3D pose data set would enable fine-tuning of the models proposed, further increasing the accuracy of the Pose Estimation process. 

We have shown a path that can address multiple problems in biomechanics, which can start taking advantage of big data. Single athlete sports, like high jump or discus throwing, can take advantage of the methods to automatically analyze both biomechanics (jumping movement or rotation speed) and actual performance information (like jumping or release angle). In other sports, like soccer or boxing, we can use these techniques to improve set piece taking or measure training performance in specific drills. 

\section{Conclusion}

We reviewed the literature related to Computer Vision methods in a simplified manner to introduce the concepts used in the field to incentive their use in sports-related research. Specifically, for biomechanical analysis, we think that the existing techniques can revolutionize the way we conduct research. 

This article's primary conclusion is that these methods can already produce gigabyte-scale data sets, which are orders of magnitude greater than current data sets. Moreover, with increasing magnitude, the associations found in data will have increased relevance, leading to more and better insights into how athletes perform and improve performance. 

With applications spanning from sports to ergonomics, or even the metaverse, these methods will play a prominent role in the near future. 

\subsection{Future Work}
There are many research paths that these solutions open. The next step is to improve the models developed in this article to achieve a higher correlation. Furthermore, we want to introduce interpretable methods and understand what insights and recommendations we can perform from this kind of data. 
\section*{Acknowledgement(s)}


This work is financed by National Funds through the Portuguese funding agency, FCT - Fundação para a Ciência e a Tecnologia, within project UIDB/50014/2020.

The second author would like to thank Futebol Clube Porto – Futebol SAD for the funding.
Any opinions, findings, and conclusions or recommendations expressed in this material are those of the authors and do not necessarily reflect the views of the cited entity.





\bibliographystyle{elsarticle-num}
\bibliography{ecrc-template}







\end{document}